\documentclass[12pt]{article}
\usepackage[utf8]{inputenc}
\usepackage[english]{babel}
\usepackage[T1]{fontenc}
\usepackage{hyperref}
\usepackage{url}
\usepackage{booktabs}
\usepackage{amsfonts}
\usepackage{nicefrac}
\usepackage{microtype}
\usepackage{lipsum}
\usepackage{graphicx}
\usepackage{natbib}
\usepackage{todonotes}
\usepackage{algpseudocode}
\usepackage{algorithm}
\usepackage{amsmath}
\usepackage{authblk}
\usepackage{multirow}
\usepackage{mdframed}
\usepackage{tikz}
\usepackage{amsmath, amssymb, bm}
\usepackage{booktabs}
\usepackage{tabularx} 
\usepackage{multirow}
\usepackage{longtable}
\usetikzlibrary{tikzmark}
\graphicspath{{media/}}
\usepackage[margin=1in]{geometry}
\providecommand{\keywords}[1]{\textbf{\textit{Index terms---}} #1}
\usepackage{caption} 
\begin{document}
\title{Attention-based Reinforcement Learning for Combinatorial Optimization: Application to Job Shop Scheduling Problem}
\author[1]{Jaejin Lee}
\author[2]{Seho Kee}
\author[2]{Mani Janakiram}
\author[1]{George Runger}
\affil[1]{Arizona State University}
\affil[2]{Intel Corporation}

\maketitle
\begin{abstract}
      Job shop scheduling problems represent a significant and complex facet of combinatorial optimization problems, which have traditionally been addressed through either exact or approximate solution methodologies. However, the practical application of these solutions is often challenged due to the complexity of real-world problems. Even when utilizing an approximate solution approach, the time required to identify a near-optimal solution can be prohibitively extensive, and the solutions derived are generally not applicable to new problems. This study proposes an innovative attention-based reinforcement learning method specifically designed for the category of job shop scheduling problems. This method integrates a policy gradient reinforcement learning approach with a modified transformer architecture. A key finding of this research is the ability of our trained learners within the proposed method to be repurposed for larger-scale problems that were not part of the initial training set. Furthermore, empirical evidence demonstrates that our approach surpasses the results of recent studies and outperforms commonly implemented heuristic rules. This suggests that our method offers a promising avenue for future research and practical application in the field of job shop scheduling problems.
\end{abstract}

\section{Introduction}
Job shop scheduling problem (JSSP) is a well-known and one of the hardest classes of combinatorial optimization problems in operations research, computer science, and industrial engineering. It has a long history as a research topic and is still being actively studied because of its applicability in numerous industrial areas, such as manufacturing and production domains \citep{gao2020solving}. 

JSSP has been tackled in mainly two ways, by exact solution approaches and approximate approaches \citep{zhang2019review, kwon2021matrix, xie2019review}. Exact solution approaches with mathematical programming have been widely adopted \citep{ku2016mixed, pei2020column, wang2018two, zhang2018flexible}; however, it may take a prohibitive amount of time to find an optimal solution, or the problems can be infeasible to find an optimal solution. This can be problematic for real-world JSSP, where a large number of jobs or machines would be associated. This has led researchers and practitioners to turn their attention to approximate methods, which aim to achieve near-optimal solutions instead. Various heuristic approaches have been applied to solve JSSP~\citep{gonccalves2005hybrid, park2003hybrid, peng2015tabu, kawaguchi2016reactive, chakraborty2015efficient, cruz2017accelerated, nagata2018guided, shi2020hybrid}.

Although the above mentioned approaches have gained success in addressing JSSP and are recognized as effective means to solve this particular problem type, solutions by those methods are subject to the constraints that are assumed when solving the problems. For example, solutions using the above methods are for the particular problems where the methods are applied, and thus, the solutions are not applicable to other problems in general. New problems are solved anew even with the same approach. Considering the shortcomings of classical solution approaches, it is essential to have a framework that can effectively solve large-scale problems without needing model adjustments or human intervention.

To address those challenges, we propose a learning method that is capable of providing solutions to unseen problems. In particular, we employ attention mechanisms \citep{bahdanau2014neural} and integrate them with reinforcement learning \citep{sutton2018reinforcement} and the transformer architecture \citep{vaswani2017attention}. We show that our trained learners can be reused to solve problems not only of the same size but also of larger sizes without retraining and demonstrate that our proposed approach outperforms the results in recent studies and widely adopted heuristic rules.

Building on this foundation, we highlight the principal contributions of our study:
\begin{itemize}
  \item The suggested model introduces a novel approach for representing the characteristics operations in JSSP by utilizing a revised transformer architecture, two independent transformer encoders attending operations in same job groups and machine sharing operations, respectively.
  \item The proposed model is agnostic to problem size. A trained policy in simpler scenarios, like 6 jobs and 6 machines, can effectively solve unseen, more complex problem instances than used in training.    
   \item A trained policy with the proposed model outperforms popularly used dispatching heuristics and other proposed models employing RL frameworks.
   \item We conducted extensive testing on synthetic and the seven most recognized benchmark instances of varying complexities to demonstrate the proposed model's broad applicability and performance.
\end{itemize}

The remainder of the paper proceeds as follows. Related work is provided in Section \ref{sec: related}, the problem is defined in Section \ref{sec: prob_def}, details of our methodologies are proposed in Section \ref{sec: method}, the experiments are described, and the results are provided in Section \ref{sec: results}, and the study is concluded in Section \ref{sec: conclusions}.
\section{Related Work}    \label{sec: related}
In recent years, the application of machine learning and deep learning methods has expanded to the field of combinatorial optimization problems (COPs), providing innovative approaches to tackle problems that have traditionally been considered challenging due to their complexity and computational demands.~\cite{vinyals2015pointer} proposed the Pointer Network, a sequence-to-sequence model to generate a permutation of the input sequence instead of a sequence from a fixed-size dictionary. This allows the model to select and order elements from the input sequence. In the original work of~\cite{vinyals2015pointer}, the traveling salesman problem (TSP) was studied. However, the model is trained with supervised learning and optimal solutions obtained from alternative methods such as the Held-Karp~\citep{held1962dynamic} algorithm for TSP. The study introduced the potential for implementing a learning-based approach in addressing COPs. A critical challenge, however, arises from its dependency on supervised learning, which requires optimal solutions for model training, but it is not possible to get optimal solutions for complex instances in most cases.

To address this issue,~\cite{bello2017neural} introduced a reinforcement learning (RL) framework as an alternative to supervised learning to train models. Like~\cite{vinyals2015pointer}, they also employed the Pointer Network architecture and studied TSPs in their research. The RL framework trains the model to learn an effective policy via a trial-and-error process, interacting with the problem environment to gradually improve its solution-finding capabilities. Unlike supervised learning used in~\cite{vinyals2015pointer}'s approach, it does not require an optimal solution for model training, making it a practical approach for solving COPs. 

\cite{kool2018attention} applied a transformer architecture \citep{vaswani2017attention} with a transformer encoder and a point network type of decoder to solve TSPs and vehicle routing problems (VRPs). The proposed model, leveraging the Transformer architecture, captures the complex inter-relationships among inputs more effectively than the model by~\cite{bello2017neural}, which employs Long Short-Term Memory (LSTM)~\citep{hochreiter1997long}. This advantage is attributed to the Transformer's superior ability to represent long-range dependencies \citep{vaswani2017attention}.
\cite{nazari2018reinforcement} proposed a modified encoder architecture by replacing the LSTM unit with 1D convolutional layers to have a model that is invariant to the input sequence order, handling the dynamic state changes to solve extended the VRPs. Subsequent studies followed the RL framework with different architectures and training methods to solve combinatorial optimization COPs, mainly for TSPs and VRPs \citep{miki2018applying, deudon2018learning, ma2019combinatorial, kwon2020pomo, xu2021reinforcement}.

Beginning with well-known TSPs and VRPs, research efforts have extended to JSSP with a similar framework. \cite{zhang2020learning} proposed a model that uses deep reinforcement learning to prioritize an operation from a set of possible operations. They used the Graph Neural Network (GNN) to represent operations' states in a disjunctive graph. One advantage of using GNN to represent states is that the proposed model is agnostic to the size of the problem, i.e., the model will not be dependent on the number of jobs and machines of the problem instances. However, a drawback of their model lies in the reward function's reliance on an estimated makespan instead of the actual makespan, potentially degrading the model's performance.


Numerous studies~\cite{park2021schedulenet, park2021learning, chen2022deep, yuan2023solving} have adopted graph neural networks or variants to embed operations similarly to \cite{zhang2020learning} with different reward functions, training algorithms, masking technique, and model structures. A motivation for using GNN is to represent the inter-relationships of operations in a disjunctive graph effectively. In a disjunctive graph, each operation is characterized by two main attributes: its relationship within a specific job group (via conjunctive edges) and its use of shared machines (via disjunctive edges). Potentially, keeping these attributes separate would help accurately represent the unique characteristics of each operation. However, a disjunctive graph is connected by both conjunctive and disjunctive edges, which results in a mixture of these important details.

 
Apart from using a graph neural network, \cite{tassel2021reinforcement} proposed an RL framework with seven designed scalar variables to capture the current state of individual jobs. However, their framework is subject to the problem size and requires retraining or adjustments to the model to solve new problems of different sizes. 

\cite{liu2020actor} introduced another RL framework based on a convolutional neural network architecture for feature extraction. One notable drawback of the proposed model is employing too simple action space, selecting the next operation by three different dispatching rules such as first-in-first-out (FIFO), shortest processing time first (SPT), and longest processing time first (LPT). These simplistic action spaces could restrict candidates for the operations that can be scheduled and potentially degrade the quality of the solutions. 
\section{Problem Definition}
\label{sec: prob_def}
A JSSP contains a set of $J$ jobs, for $j=1, 2, \ldots, J$, which are to be processed in a shop that consists of $M$ machines for $m=1, 2,\ldots, M$. 
Each job is comprised of $M$ operations, $O_{jm}$ for for $j=1, 2, \ldots, J$ and $m=1, 2, \ldots, M$, where each operation $O_{jm}$ can be processed on specific machine $m$. 
Additionally, each operation $O_{jm}$ has a predefined order $t$ in the sequence of operations. That is, $O^{t}_{jm}$ can be processed if and only if all the prior operations, $O^{t'}_{jm}$, where $t' < t$, have been completed. One machine processes exactly one operation at a time without interruption. The time to complete an operation, $O_{jm}$, is called the processing time, denoted as $p_{jm}$. Completion of all the element operations completes a job. A valid sequence of job assignments to machines generates a feasible schedule. The objective is to learn a schedule or a scheduling policy that optimizes a function of feasible schedules, e.g., the minimization of makespan or the maximization of machine utilization.

\section{Attention-based Reinforcement Learning for JSSP}
\label{sec: method}
We propose a transformer-based reinforcement learning scheduler (ARLS) with enhancements on top of the original transformer architecture~\citep{vaswani2017attention}. ARLS specifically aims to capture the characteristics of operations better, focusing on job groups and machine-sharing aspects using two independent transformer encoders.

The input vector of each operation is encoded by linear projection, which is then transformed into embedding by the encoder using a self-attention mechanism. Two independent encoders produce two different operation embeddings from the same initial encoding. The decoder applies encoder-decoder attention using embeddings from both encoders and the decoder input. More details about this will be provided in subsequent sections. 



\subsection{Input}
    The input vectors $\mathbf{o}_n$, for $n=1, 2, ..., JM+1$, are concatenated from four vectors: 1) job index, 2) operation order in the job sequence, 3) machine index of the operation, and 4) processing time of the operation. 
    The operation index $n = JM+1$ accounts for the cases when no operation is assigned. i.e., the machine is left idle.
    
    The operations are initially encoded via the linear projection of the input as shown in equation \ref{initial_embedding}.
      \begin{equation}
            \bm{h}^{(0)} = \mathbf{W}^{(0)} \bm{o}_n + \bm{b}^{(0)}
            \label{initial_embedding}
        \end{equation}
    where $\bm{o}_n$, for $n=1, 2, ..., JM+1$, are the input vectors, $\mathbf{W}^{(0)}$ is the weight matrix, and $\bm{b}^{(0)}$ is the bias vector. 

\subsection{Encoder}
   We have modified the original transformer encoder to effectively represent the characteristics of operations. In contrast to the traditional transformer encoder, ARLS has two separate encoders that both take the same initial input encoding, denoted as $\bm{h}^{(0)}$ as input. The input encodings are transformed from the initial encoding $\bm{h}^{(0)}$ to the embedding $\bm{h}^{(L)}$ through multiple, e.g., $L$, encoder layers. 
   
  The original transformer employs simple masked attention in its encoder layers to prevent the model from attending to padding, which is used to equalize the lengths of input tokens in natural language processing (NLP). In contrast to NLP, COPs have inputs of identical length, eliminating the need for padding. Instead, ARLS employs two distinct attention mechanisms in each encoder layer. These mechanisms are designed to focus attention specifically on operations within the same job group and operations sharing machines, respectively.


Consider the scaled dot-product attention function    
    \begin{equation}
        \text{Attention}(Q, K, V) = \text{softmax} \left( \frac{QK^{\intercal}}{\sqrt{d_K}}\right)V
        \label{eq: attention}
    \end{equation}
    where $Q, K, V$ are each head's query, key, and value matrices, respectively, and $d_K$ is the dimension for each head's query and key matrices. 
    Let each element $c_{i,j} = \bm{q}_i \bm{k}^{T}_{j}/\sqrt{d_K}$ where $\bm{q}$, $\bm{k}$ are $i$th and $j$th columns of the query and key matrices, $Q$ and $K$, respectively.
   
    We apply two different masking schemes to $c_{i,j}$, one for the jobs, denoted as $c_{i,j}^{\mathcal{(J)}}$, and another for the machines, $c_{i,j}^{\mathcal{(M)}}$, and integrate the two encoder results at the end of the encoder. In particular, the values of the elements $c_{i,j}^{\mathcal{(J)}} = -\infty$ are masked if operations $i$ and $j$ are not in the same job sequence, and similarly, $c_{i,j}^{\mathcal{(M)}} = -\infty$ if operations $i$ and $j$ are not processed by the same machine.

    
The embedding of operations from the last layer of the encoder $\bm{h}^{(L)}$ is computed as a convex combination of the two embeddings \(\bm{h}^{(L)}_{\mathcal{J}}\) and \(\bm{h}^{(L)}_{\mathcal{M}}\) as
    \begin{equation}
        \bm{h}^{(L)} = \lambda\bm{h}^{(L)}_{\mathcal{J}} + (1-\lambda)\bm{h}^{(L)}_{\mathcal{M}}
        \label{final_embedding}
    \end{equation}
where $0 \le \lambda \le 1$ controls the importance of each embedding. 

\subsection{Decoder}
The decoder has $L^*$ identical attention layers similar to the original transformer model~\citep{vaswani2017attention}. However, masking is performed respective to the operations that were already assigned. Let $\mathcal{O}_{t}^{-} \subset \mathcal{O}$ be the set of operations that were already assigned by time $t$. The complement is $\mathcal{O}_{t}^{+} = \mathcal{O} \setminus \mathcal{O}_{t}^{-}$. Also, $\bm{h}_{t-1}$ denotes the encoder embedding of the operation assigned at the previous time step $t-1$.

For decoder attention, which is also a masked, multi-head attention with scaled dot-product, we use the encoder embedding of the last assigned operation $\bm{h}_{t-1}$ for query vector, $\bm{q}$, and the encoder embeddings from the previous time step $\bm{h}^{(L)}$ for keys $K$ and values $V$ 

    An action selection module generates a probability distribution over possible operations on the top of the decoder. Attention scores are calculated using the decoder output at time $t$, denoted as $\bm{q}_{(t)}$ for query, and the previous encoder embeddings $\bm{h}^{(L)}$ for keys $K$ to generate probabilities $p_{o}$ over the possible operations $\mathcal{O}^{+}$. Nonlinear activation function $tanh$ is applied with the clipping technique~\citep{li2023deep} before the softmax function is applied.     
     \begin{align}
        p_{o} &= 
        \begin{cases} 
        	\text{softmax}\left(C\text{tanh}\left(\bm{q}_{(t)}\bm{k}_o^{\intercal} / \sqrt{d_K}\right)\right), & \text{if } \text{operation} \,\, j \notin \mathcal{O}^{-}\\
            0 & \text{otherwise}
        \end{cases}            
    \end{align}
    where $C$ is the clipping constant. 
    The probability $p_{o}$ is used to choose operation at time step $t$.

\subsection{Training}
We modify the Monte-Carlo policy gradient method~\citep{sutton2018reinforcement, bahdanau2016actor} for training. The performance of learned
policy was estimated by a Monte-Carlo method using observed makespans. The average of the makespans respective to the same problem is used as the baseline. Observed makespans minus the baseline generate the advantages used for gradient updates 
\begin{equation}
        \nabla_{\theta} J(\pi_{\theta}) = \underset{\tau \sim \pi_{\theta}}{\mathrm{E}} \left[\sum_{t=0}^{T} \nabla_{\theta} \ln \pi_{\theta} (a_t|s_t) \left( R(\tau) - b \right) \right]
\label{montecarlo}        
\end{equation}
where $\tau$ is a trajectory of an episode, $R(\tau)$ is the Monte-Carlo episodic reward, i.e., makespan of a trajectory, and $b$ is a baseline to reduce variance. 

Preferred by researchers, as cited in works by~\citep{mnih2013playing, schulman2015high, haarnoja2018soft, reda2020learning}, the baseline $b$ is typically derived from the value network. Unlike the typical RL model, our proposed model does not have a value network but takes a different strategy compared to the typical Monte-Carlo Policy Gradient, which generates multiple trajectories. Our multi-trajectory strategy offers several advantages compared to employing a value network. More problem instances are expected to increase the scheduling performance robustness. Furthermore, for each problem instance, the learner is expected to improve with more diverse scheduling solutions. 

Let $\mathcal{T}$ denote the set of all trajectories, that is, all the feasible solutions for a JSSP instance. Let each $\tau_n \in \mathcal{T}$ be a sampled trajectory, and $\mathcal{T}_N = \{ \tau_1, \ldots, \tau_N \}$ be the set of $N$ sampled trajectories for a JSSP instance, with which we make minibatch updates for the weights throughout the training as follows.

\begin{equation}
       \nabla_{\theta} J(\pi_{\theta}) = \frac{1}{N} \sum_{n=1}^{N}  \left(\underset{\tau \sim \pi_{\theta}}{\mathrm{E}} \left[\sum_{t=0}^{T} \nabla_{\theta} \ln \pi_{\theta} (a_t|s_t) \left( R(\tau_n) - b \right) \right] \right)
    \label{parallel_episode}
    \end{equation}
    where $R(\tau_{n})$ is the Monte-Carlo episodic reward of trajectory $n$, and $b = \frac{1}{N} \sum_{n=1}^{N}R(\tau_{n})$, which is the average of the $N$ episodic rewards obtained from the same problem instance in parallel.
\section{Experimental Results}
\label{sec: results}
The training and testing approach for ARLS prioritizes rapid model learning in simplified settings. A key objective is a learner design that effectively scales to greater problem instance sizes and maintains high performance. ARLS is trained with synthetic data with 6 jobs and 6 machines. In total 36 operations, with processing times randomly selected uniformly in the range between 1 and 15, similar to the prior study~\cite{zhang2020learning}. The Adam optimization \citep{kingma2014adam} was employed for weight updates. ARLS uses a relatively low complexity transformer architecture with only three encoder layers and a single decoder layer, the model dimension is 256, the number of heads is 16, and the fully connected feed-forward neural networks dimension is 512. Further design studies might improve performance.

Training generated 100,000 replicate synthetic 6 jobs $\times$ 6 machines problem instances. A mini-batch update~\citep{yang2019mini} is applied. The objective is to minimize the schedule's makespan. To evaluate ARLS performance, the average optimality gap is used
\begin{equation*}
	\textit{opt}_{\textit{gap}} = \frac{\textit{obj} - \textit{obj}^{*}}{\textit{obj}^{*}}
\end{equation*}
where \textit{obj} is the makespan obtained from the ARLS, and $\textit{obj}^{*}$ is either the optimal or best-known solution. For synthetic instances, the optimal solutions are derived using constraint programming with Cplex~\citep{cplex}. 

%
%

The generalized performance of ARLS is tested with varying sizes of synthetic and benchmark instances, including problem instance sizes greater than seen in training. To evaluate, we use three synthetic datasets with sizes configured as follows: the first dataset has 6 jobs and 6 machines, the second dataset consists of 10 jobs and 10 machines, and the third dataset comprises 15 jobs and 15 machines. Additionally, seven well-known benchmark sets from \cite{bench} are tested for comparison. 

Table \ref{tab:result_synthetic} shows the test results on synthetic data. The average optimality gap is obtained from 1,000 instances for each problem size. To the best of our knowledge, the comparable test on synthetic instances is the study by~\cite{zhang2020learning}(Z20). Our model is trained on only 6 jobs and 6 machine settings. In contrast, Zhang et al.,~\citep{zhang2020learning} trained their model separately with samples from the three problem sizes. On average, ARLS outperforms their results by a substantial margin of 12.5\%.
%
%

\begin{table}[h]
	\centering
	\caption{Performance of our ARLS model compared with Z20~\citep{zhang2020learning} on three different synthetic datasets. Performance is evaluated by the optimality gap in percentage}
	\begin{tabular}{cccccc}
		\toprule
		Jobs & Machines & Instances & ARLS (Ours) & Z20\\
		\midrule
		6 & 6 & 1000& 4.8 & 17.7  \\
		10 & 10 & 1000 & 10.9 & 22.3  \\
		15 & 15 & 1000 & 13.5 & 26.7 \\
		\bottomrule
	\end{tabular}    
	\label{tab:result_synthetic}
\end{table}

ARLS was tested with seven benchmark datasets of varying sizes in the number of jobs (ranging from 6 to 100) and machines (ranging from 5 to 20). Table \ref{tab: result_benchmark} shows ARLS's performance compared to work by \cite{zhang2020learning} denoted as Z20, \cite{park2021schedulenet} as P21a, \cite{park2021learning} as P21b, \cite{chen2022deep} as C22, and \cite{yuan2023solving} and Y23, and popularly adopted heuristic rules, such as first-in-first-out (FIFO), shortest processing time (SPT), and most work remaining (MWKR). Here, instead of the optimality gap, the best-known gap, which is the relative difference to the best-known makespan, is used as optimal makespans are unknown for some of the problems. Problems with multiple instances show average results.
Our ARLS model showed the best performance on all but 4 problem sizes out of 26 problem classes despite being trained in the simplest setting of 6 jobs and 6 machines. 



    \begin{table}[H]
	\centering
	\caption{Performance of our ARLS model compared with heuristic rules and other studies on benchmark datasets. Performance is evaluated by the best known gap in percentage (with averages for multiple instances). Results by \cite{zhang2020learning} denoted as Z20, \cite{park2021schedulenet} as P21a, \cite{park2021learning} as P21b, \cite{chen2022deep} as C22, and \cite{yuan2023solving} and Y23, and popularly adopted heuristic rules, such as first-in-first-out (FIFO), shortest processing time (SPT), and most work remaining (MWKR).}
	\label{tab: result_benchmark}
	\resizebox{\textwidth}{!}{%
		\begin{tabular}{crrrrrrrrrrrr}
			\hline
			\multicolumn{1}{c}{Dataset} & \multicolumn{1}{c}{Jobs} & \multicolumn{1}{c}{Machines} & \multicolumn{1}{c}{Instances} & \multicolumn{1}{c}{FIFO} & \multicolumn{1}{c}{SPT} & \multicolumn{1}{c}{MWKR} & \multicolumn{1}{c}{Z20} & \multicolumn{1}{c}{P21a} & \multicolumn{1}{c}{P21b} & \multicolumn{1}{c}{C22} & \multicolumn{1}{c}{Y23} & \multicolumn{1}{c}{\textbf{Ours}} \\ \hline
			\multirow{2}{*}{ABZ} & 10 & 10 & 2 & 14.8 & 12.9 & 9.5 & - & 6.2 & 10.1 & - & 7.1 & \textbf{6.1} \\
			& 15 & 15 & 3 & 31.3 & 33.3 & 21.3 & - & 20.5 & 29.0 & - & 22.3 & \textbf{17.6} \\ \hline
			\multirow{3}{*}{FT} & 6 & 6 & 1 & 18.2 & 60.0 & 10.9 & - & \textbf{7.3} & 29.1 & - & 9.1 & \textbf{7.3} \\
			& 15 & 15 & 1 & 27.3 & 15.5 & 24.9 & - & 19.5 & 22.8 & - & 18.5 & \textbf{11.3} \\
			& 20 & 5 & 1 & 41.2 & \textbf{8.8} & 34.8 & - & 28.7 & 14.8 & - & 14.3 & 10.1 \\ \hline
			YN & 20 & 20 & 4 & 25.9 & 30.6 & 20.8 & - & 18.4 & 24.8 & - & 20.3 & \textbf{16.1} \\ \hline
			\multirow{3}{*}{SWV} & 20 & 10 & 5 & 44.4 & 26.3 & 38.7 & - & 34.4 & 28.4 & - & 24.8 & \textbf{21.7} \\
			& 20 & 15 & 5 & 44.9 & 32.0 & 36.7 & - & 30.5 & 29.4 & - & 28.3 & \textbf{21.3} \\
			& 50 & 10 & 10 & 30.0 & 17.0 & 25.6 & - & 25.3 & 16.8 & - & 14.1 & \textbf{10.3} \\ \hline
			ORB & 10 & 10 & 10 & 29.7 & 26.3 & 29.9 & - & 20.0 & 21.8 & - & 20.9 & \textbf{13.5} \\ \hline
			\multirow{8}{*}{TA} & 15 & 15 & 10 & 21.2 & 25.9 & 23.1 & 26.0 & 15.3 & 20.1 & 35.4 & 17.9 & \textbf{13.1} \\
			& 20 & 15 & 10 & 22.8 & 32.8 & 30.0 & 30.0 & 19.5 & 25.0 & 32.1 & 20.6 & \textbf{17.9} \\
			& 20 & 20 & 10 & 24.0 & 27.8 & 27.7 & 31.6 & 17.2 & 29.3 & 28.3 & 19.5 & \textbf{16.1} \\
			& 30 & 15 & 10 & 24.0 & 35.3 & 30.2 & 33.0 & \textbf{19.1} & 24.7 & 36.4 & 22.4 & 19.6 \\
			& 30 & 20 & 10 & 25.7 & 34.4 & 30.9 & 33.6 & 23.7 & 32.0 & 34.7 & 23.3 & \textbf{22.8} \\
			& 50 & 15 & 10 & 17.8 & 24.1 & 20.1 & 22.4 & 13.9 & 15.9 & 31.9 & 16.0 & \textbf{12.9} \\
			& 50 & 20 & 10 & 18.4 & 25.2 & 23.2 & 26.5 & \textbf{13.5} & 21.3 & 28.0 & 17.4 & 16.8 \\
			& 100 & 20 & 10 & 8.8 & 14.4 & 12.8 & 13.6 & \textbf{6.7} & 9.2 & 18.0 & 8.9 & 9.3 \\ \hline
			\multirow{8}{*}{DMU} & 20 & 15 & 10 & 30.5 & 64.1 & 37.2 & 39.0 & - & - & - & 26.6 & \textbf{18.9} \\
			& 20 & 20 & 10 & 26.4 & 64.6 & 32.4 & 37.7 & - & - & - & 23.6 & \textbf{17.3} \\
			& 30 & 15 & 10 & 34.8 & 62.6 & 39.3 & 41.9 & - & - & - & 28.7 & \textbf{20.7} \\
			& 30 & 20 & 10 & 32.2 & 65.9 & 36.6 & 39.5 & - & - & - & 29.2 & \textbf{22.6} \\
			& 40 & 15 & 10 & 31.2 & 55.9 & 35.1 & 35.4 & - & - & - & 24.5 & \textbf{20.2} \\
			& 40 & 20 & 10 & 33.2 & 63.0 & 39.7 & 39.4 & - & - & - & 30.1 & \textbf{24.8} \\
			& 50 & 15 & 10 & 31.0 & 50.4 & 34.7 & 36.2 & - & - & - & 24.9 & \textbf{18.8} \\
			& 50 & 20 & 10 & 35.3 & 62.2 & 41.4 & 38.4 & - & - & - & 29.5 & \textbf{21.8} \\ \hline
		\end{tabular}%
	}	
\end{table}

%

\begin{figure}
  \centering
  \includegraphics[width=0.8\linewidth, height=0.3\textheight]{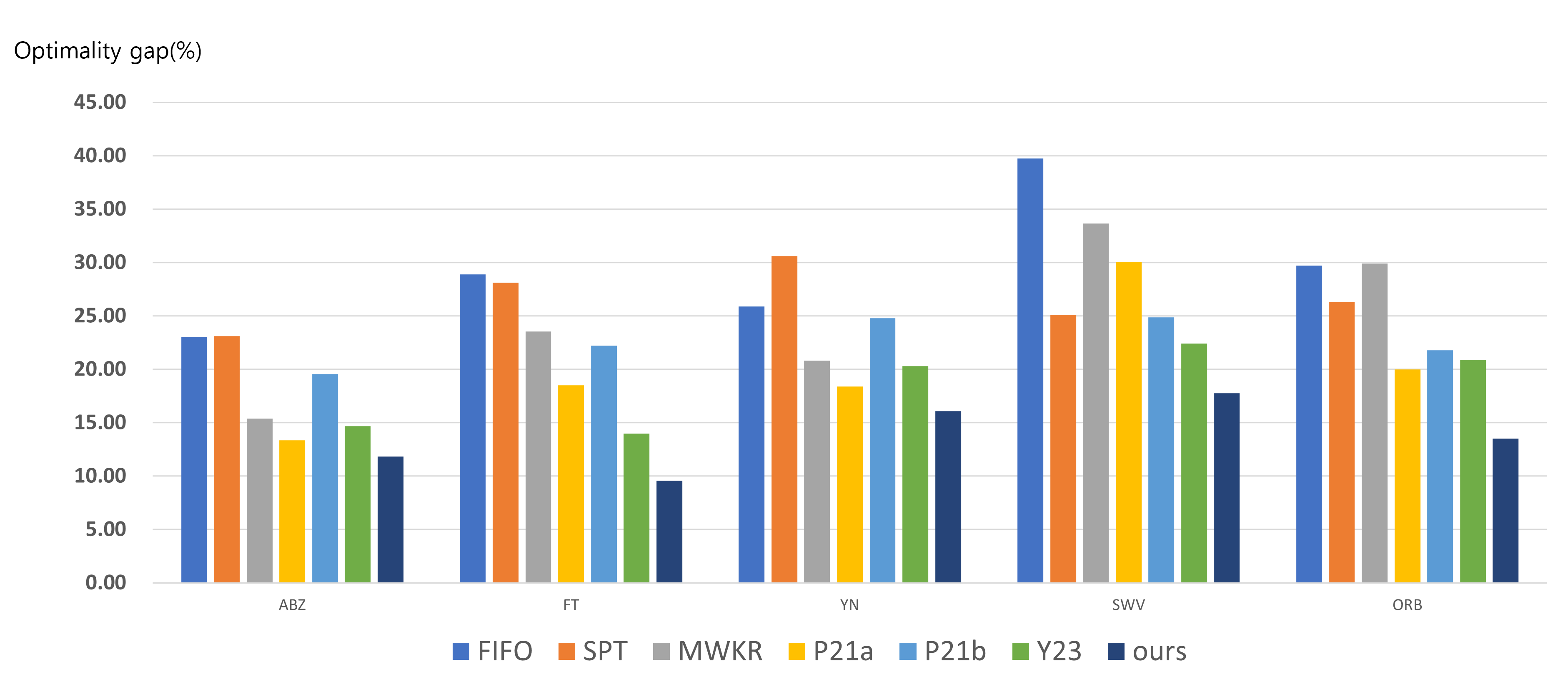}
  \caption{Generalized Performance of our ARLS model compared with heuristic rules and other studies on benchmark datasets, ABZ, FT, YN, SWVM, and ORB.
  Results by \cite{zhang2020learning} denoted as Z20, \cite{park2021schedulenet} as P21a, \cite{park2021learning} as P21b and \cite{yuan2023solving} and Y23, and popularly adopted heuristic rules, such as first-in-first-out (FIFO), shortest processing time (SPT), and most work remaining (MWKR).}
  \label{fig:result_1}
\end{figure}  

\begin{figure}
	\centering
	\includegraphics[width=0.8\linewidth, height=0.3\textheight]{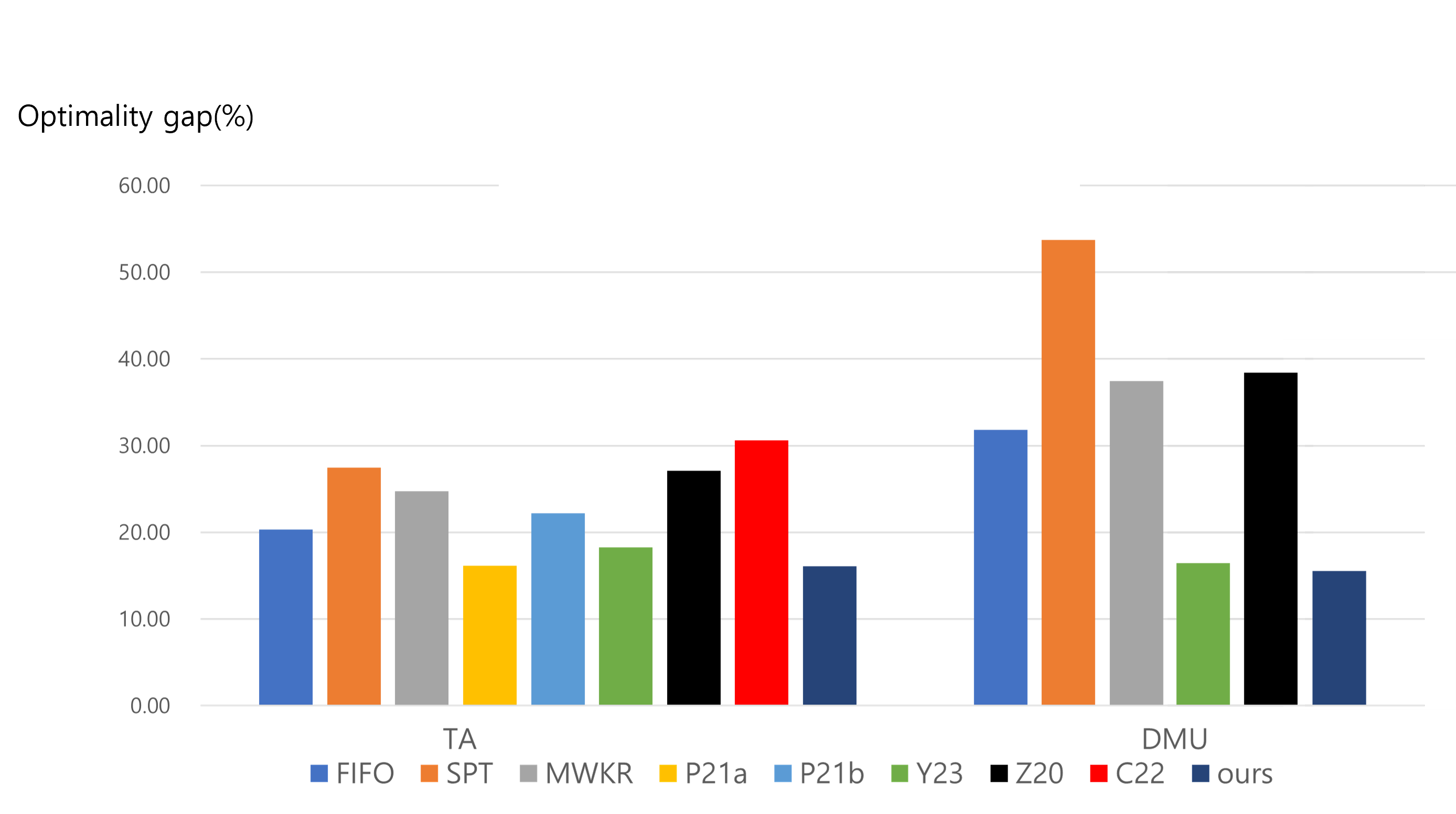}
	\caption{Generalized Performance of our ARLS model compared with heuristic rules and other studies on benchmark datasets, TA and DMU.
  Results by \cite{zhang2020learning} denoted as Z20, \cite{chen2022deep} as C22, and \cite{yuan2023solving} and Y23, and popularly adopted heuristic rules, such as first-in-first-out (FIFO), shortest processing time (SPT), and most work remaining (MWKR).}
	\label{fig:result_2}
\end{figure}

On average, ARLS outperformed other algorithms in all cases, as shown in Figures~\ref{fig:result_1} and \ref{fig:result_2}. ARLS was exceeded only in a few problem classes in TA. Even then, the P21a algorithm was trained on problem instances about 67\% greater than the training size for ARLS. 

Although ARLS was exceeded by P21a for three TA classes: TA(30, 15), TA(50, 20), and TA(100, 20) when ARLS trained in the 6 jobs and 6 machines scenario, ARLS outperformed P21a when trained on datasets of the equivalent size of P21a, resulting smaller optimality gaps, with 16.8\% vs. 19.1\% for TA(30,15), 9.96\% vs. 13..5\% for TA(50,20), and 4.4\% vs. 6.7\% for TA(100,20), respectively.

\section{Conclusions}    \label{sec: conclusions}
We developed an attention-based reinforcement learner for job shop scheduling problems with a modified transformer architecture. A key point of our model is that it enables the learner to scale with problem instance size. The learner was trained with synthetic problem instances of relatively small size in the number of jobs and machines and scaled to new problems of various larger sizes with both synthetic and benchmark datasets. We have obtained generally outperforming results compared to recent studies and widely adopted heuristic rules. The model architecture is new, but the transformer is low in complexity (few layers), and further work might improve results with larger transformer architectures. 


\bibliographystyle{apalike}
\bibliography{references}  

\begin{thebibliography}{}

\bibitem[Bahdanau et~al., 2016]{bahdanau2016actor}
Bahdanau, D., Brakel, P., Xu, K., Goyal, A., Lowe, R., Pineau, J., Courville, A., and Bengio, Y. (2016).
\newblock An actor-critic algorithm for sequence prediction.
\newblock {\em arXiv preprint arXiv:1607.07086}.

\bibitem[Bahdanau et~al., 2014]{bahdanau2014neural}
Bahdanau, D., Cho, K., and Bengio, Y. (2014).
\newblock Neural machine translation by jointly learning to align and translate.
\newblock {\em arXiv preprint arXiv:1409.0473}.

\bibitem[Bello et~al., 2017]{bello2017neural}
Bello, I., Pham, H., Le, Q.~V., Norouzi, M., and Bengio, S. (2017).
\newblock Neural combinatorial optimization with reinforcement learning.

\bibitem[Chakraborty and Bhowmik, 2015]{chakraborty2015efficient}
Chakraborty, S. and Bhowmik, S. (2015).
\newblock An efficient approach to job shop scheduling problem using simulated annealing.
\newblock {\em International Journal of Hybrid Information Technology}, 8(11):273--284.

\bibitem[Chen et~al., 2022]{chen2022deep}
Chen, R., Li, W., and Yang, H. (2022).
\newblock A deep reinforcement learning framework based on an attention mechanism and disjunctive graph embedding for the job-shop scheduling problem.
\newblock {\em IEEE Transactions on Industrial Informatics}, 19(2):1322--1331.

\bibitem[Cruz-Ch{\'a}vez et~al., 2017]{cruz2017accelerated}
Cruz-Ch{\'a}vez, M.~A., Mart{\'\i}nez-Rangel, M.~G., and Cruz-Rosales, M.~H. (2017).
\newblock Accelerated simulated annealing algorithm applied to the flexible job shop scheduling problem.
\newblock {\em International Transactions in Operational Research}, 24(5):1119--1137.

\bibitem[Deudon et~al., 2018]{deudon2018learning}
Deudon, M., Cournut, P., Lacoste, A., Adulyasak, Y., and Rousseau, L.-M. (2018).
\newblock Learning heuristics for the tsp by policy gradient.
\newblock In {\em Integration of Constraint Programming, Artificial Intelligence, and Operations Research: 15th International Conference, CPAIOR 2018, Delft, The Netherlands, June 26--29, 2018, Proceedings 15}, pages 170--181. Springer.

\bibitem[Gao et~al., 2020]{gao2020solving}
Gao, D., Wang, G.-G., and Pedrycz, W. (2020).
\newblock Solving fuzzy job-shop scheduling problem using de algorithm improved by a selection mechanism.
\newblock {\em IEEE Transactions on Fuzzy Systems}, 28(12):3265--3275.

\bibitem[Gon{\c{c}}alves et~al., 2005]{gonccalves2005hybrid}
Gon{\c{c}}alves, J.~F., de~Magalh{\~a}es~Mendes, J.~J., and Resende, M.~G. (2005).
\newblock A hybrid genetic algorithm for the job shop scheduling problem.
\newblock {\em European journal of operational research}, 167(1):77--95.

\bibitem[Haarnoja et~al., 2018]{haarnoja2018soft}
Haarnoja, T., Zhou, A., Abbeel, P., and Levine, S. (2018).
\newblock Soft actor-critic: Off-policy maximum entropy deep reinforcement learning with a stochastic actor.
\newblock In {\em International conference on machine learning}, pages 1861--1870. PMLR.

\bibitem[Held and Karp, 1962]{held1962dynamic}
Held, M. and Karp, R.~M. (1962).
\newblock A dynamic programming approach to sequencing problems.
\newblock {\em Journal of the Society for Industrial and Applied mathematics}, 10(1):196--210.

\bibitem[Hochreiter and Schmidhuber, 1997]{hochreiter1997long}
Hochreiter, S. and Schmidhuber, J. (1997).
\newblock Long short-term memory.
\newblock {\em Neural computation}, 9(8):1735--1780.

\bibitem[{IBM Decision Optimization}, 2024]{cplex}
{IBM Decision Optimization} (Accessed 2024).
\newblock Constraint programming modeling for python (docplex.cp).
\newblock DOcplex.CP: Constraint Programming Modeling for Python V2.25 documentation.

\bibitem[Kawaguchi and Fukuyama, 2016]{kawaguchi2016reactive}
Kawaguchi, S. and Fukuyama, Y. (2016).
\newblock Reactive tabu search for job-shop scheduling problems.
\newblock In {\em 2016 11th International Conference on Computer Science \& Education (ICCSE)}, pages 97--102. IEEE.

\bibitem[Kingma and Ba, 2014]{kingma2014adam}
Kingma, D.~P. and Ba, J. (2014).
\newblock Adam: A method for stochastic optimization.
\newblock {\em arXiv preprint arXiv:1412.6980}.

\bibitem[Kool et~al., 2018]{kool2018attention}
Kool, W., Van~Hoof, H., and Welling, M. (2018).
\newblock Attention, learn to solve routing problems!
\newblock {\em arXiv preprint arXiv:1803.08475}.

\bibitem[Ku and Beck, 2016]{ku2016mixed}
Ku, W.-Y. and Beck, J.~C. (2016).
\newblock Mixed integer programming models for job shop scheduling: A computational analysis.
\newblock {\em Computers \& Operations Research}, 73:165--173.

\bibitem[Kwon et~al., 2020]{kwon2020pomo}
Kwon, Y.-D., Choo, J., Kim, B., Yoon, I., Gwon, Y., and Min, S. (2020).
\newblock Pomo: Policy optimization with multiple optima for reinforcement learning.
\newblock {\em Advances in Neural Information Processing Systems}, 33:21188--21198.

\bibitem[Kwon et~al., 2021]{kwon2021matrix}
Kwon, Y.-D., Choo, J., Yoon, I., Park, M., Park, D., and Gwon, Y. (2021).
\newblock Matrix encoding networks for neural combinatorial optimization.
\newblock {\em Advances in Neural Information Processing Systems}, 34:5138--5149.

\bibitem[Li, 2023]{li2023deep}
Li, S.~E. (2023).
\newblock Deep reinforcement learning.
\newblock In {\em Reinforcement Learning for Sequential Decision and Optimal Control}, pages 365--402. Springer.

\bibitem[Liu et~al., 2020]{liu2020actor}
Liu, C.-L., Chang, C.-C., and Tseng, C.-J. (2020).
\newblock Actor-critic deep reinforcement learning for solving job shop scheduling problems.
\newblock {\em Ieee Access}, 8:71752--71762.

\bibitem[Ma et~al., 2019]{ma2019combinatorial}
Ma, Q., Ge, S., He, D., Thaker, D., and Drori, I. (2019).
\newblock Combinatorial optimization by graph pointer networks and hierarchical reinforcement learning.
\newblock {\em arXiv preprint arXiv:1911.04936}.

\bibitem[Miki et~al., 2018]{miki2018applying}
Miki, S., Yamamoto, D., and Ebara, H. (2018).
\newblock Applying deep learning and reinforcement learning to traveling salesman problem.
\newblock In {\em 2018 international conference on computing, electronics \& communications engineering (ICCECE)}, pages 65--70. IEEE.

\bibitem[Mnih et~al., 2013]{mnih2013playing}
Mnih, V., Kavukcuoglu, K., Silver, D., Graves, A., Antonoglou, I., Wierstra, D., and Riedmiller, M. (2013).
\newblock Playing atari with deep reinforcement learning.
\newblock {\em arXiv preprint arXiv:1312.5602}.

\bibitem[Nagata and Ono, 2018]{nagata2018guided}
Nagata, Y. and Ono, I. (2018).
\newblock A guided local search with iterative ejections of bottleneck operations for the job shop scheduling problem.
\newblock {\em Computers \& Operations Research}, 90:60--71.

\bibitem[Nazari et~al., 2018]{nazari2018reinforcement}
Nazari, M., Oroojlooy, A., Snyder, L., and Tak{\'a}c, M. (2018).
\newblock Reinforcement learning for solving the vehicle routing problem.
\newblock {\em Advances in neural information processing systems}, 31.

\bibitem[Park et~al., 2003]{park2003hybrid}
Park, B.~J., Choi, H.~R., and Kim, H.~S. (2003).
\newblock A hybrid genetic algorithm for the job shop scheduling problems.
\newblock {\em Computers \& industrial engineering}, 45(4):597--613.

\bibitem[Park et~al., 2021a]{park2021schedulenet}
Park, J., Bakhtiyar, S., and Park, J. (2021a).
\newblock Schedulenet: Learn to solve multi-agent scheduling problems with reinforcement learning.
\newblock {\em arXiv preprint arXiv:2106.03051}.

\bibitem[Park et~al., 2021b]{park2021learning}
Park, J., Chun, J., Kim, S.~H., Kim, Y., and Park, J. (2021b).
\newblock Learning to schedule job-shop problems: representation and policy learning using graph neural network and reinforcement learning.
\newblock {\em International Journal of Production Research}, 59(11):3360--3377.

\bibitem[Pei et~al., 2020]{pei2020column}
Pei, Z., Zhang, X., Zheng, L., and Wan, M. (2020).
\newblock A column generation-based approach for proportionate flexible two-stage no-wait job shop scheduling.
\newblock {\em International journal of production research}, 58(2):487--508.

\bibitem[Peng et~al., 2015]{peng2015tabu}
Peng, B., L{\"u}, Z., and Cheng, T. C.~E. (2015).
\newblock A tabu search/path relinking algorithm to solve the job shop scheduling problem.
\newblock {\em Computers \& Operations Research}, 53:154--164.

\bibitem[Reda et~al., 2020]{reda2020learning}
Reda, D., Tao, T., and van~de Panne, M. (2020).
\newblock Learning to locomote: Understanding how environment design matters for deep reinforcement learning.
\newblock In {\em Proceedings of the 13th ACM SIGGRAPH Conference on Motion, Interaction and Games}, pages 1--10.

\bibitem[Schulman et~al., 2015]{schulman2015high}
Schulman, J., Moritz, P., Levine, S., Jordan, M., and Abbeel, P. (2015).
\newblock High-dimensional continuous control using generalized advantage estimation.
\newblock {\em arXiv preprint arXiv:1506.02438}.

\bibitem[Shi et~al., 2020]{shi2020hybrid}
Shi, F., Zhao, S., and Meng, Y. (2020).
\newblock Hybrid algorithm based on improved extended shifting bottleneck procedure and ga for assembly job shop scheduling problem.
\newblock {\em International Journal of Production Research}, 58(9):2604--2625.

\bibitem[Shylo, 2010]{bench}
Shylo, O. (2010).
\newblock Job shop scheduling problem instances.

\bibitem[Sutton and Barto, 2018]{sutton2018reinforcement}
Sutton, R.~S. and Barto, A.~G. (2018).
\newblock {\em Reinforcement learning: An introduction}.
\newblock MIT press.

\bibitem[Tassel et~al., 2021]{tassel2021reinforcement}
Tassel, P., Gebser, M., and Schekotihin, K. (2021).
\newblock A reinforcement learning environment for job-shop scheduling.
\newblock {\em arXiv preprint arXiv:2104.03760}.

\bibitem[Vaswani et~al., 2017]{vaswani2017attention}
Vaswani, A., Shazeer, N., Parmar, N., Uszkoreit, J., Jones, L., Gomez, A.~N., Kaiser, {\L}., and Polosukhin, I. (2017).
\newblock Attention is all you need.
\newblock {\em Advances in neural information processing systems}, 30.

\bibitem[Vinyals et~al., 2015]{vinyals2015pointer}
Vinyals, O., Fortunato, M., and Jaitly, N. (2015).
\newblock Pointer networks.
\newblock {\em Advances in neural information processing systems}, 28.

\bibitem[Wang et~al., 2018]{wang2018two}
Wang, H., Jiang, Z., Wang, Y., Zhang, H., and Wang, Y. (2018).
\newblock A two-stage optimization method for energy-saving flexible job-shop scheduling based on energy dynamic characterization.
\newblock {\em Journal of Cleaner Production}, 188:575--588.

\bibitem[Xie et~al., 2019]{xie2019review}
Xie, J., Gao, L., Peng, K., Li, X., and Li, H. (2019).
\newblock Review on flexible job shop scheduling.
\newblock {\em IET Collaborative Intelligent Manufacturing}, 1(3):67--77.

\bibitem[Xu et~al., 2021]{xu2021reinforcement}
Xu, Y., Fang, M., Chen, L., Xu, G., Du, Y., and Zhang, C. (2021).
\newblock Reinforcement learning with multiple relational attention for solving vehicle routing problems.
\newblock {\em IEEE Transactions on Cybernetics}, 52(10):11107--11120.

\bibitem[Yang et~al., 2019]{yang2019mini}
Yang, Z., Wang, C., Zhang, Z., and Li, J. (2019).
\newblock Mini-batch algorithms with online step size.
\newblock {\em Knowledge-Based Systems}, 165:228--240.

\bibitem[Yuan et~al., 2023]{yuan2023solving}
Yuan, E., Cheng, S., Wang, L., Song, S., and Wu, F. (2023).
\newblock Solving job shop scheduling problems via deep reinforcement learning.
\newblock {\em Applied Soft Computing}, 143:110436.

\bibitem[Zhang et~al., 2020]{zhang2020learning}
Zhang, C., Song, W., Cao, Z., Zhang, J., Tan, P.~S., and Chi, X. (2020).
\newblock Learning to dispatch for job shop scheduling via deep reinforcement learning.
\newblock {\em Advances in Neural Information Processing Systems}, 33:1621--1632.

\bibitem[Zhang et~al., 2019]{zhang2019review}
Zhang, J., Ding, G., Zou, Y., Qin, S., and Fu, J. (2019).
\newblock Review of job shop scheduling research and its new perspectives under industry 4.0.
\newblock {\em Journal of Intelligent Manufacturing}, 30(4):1809--1830.

\bibitem[Zhang and Wang, 2018]{zhang2018flexible}
Zhang, S. and Wang, S. (2018).
\newblock Flexible assembly job-shop scheduling with sequence-dependent setup times and part sharing in a dynamic environment: Constraint programming model, mixed-integer programming model, and dispatching rules.
\newblock {\em IEEE Transactions on Engineering Management}, 65(3):487--504.

\end{thebibliography}

\end{document}